\pdfoutput=1
\documentclass[sigconf]{acmart}
\AtBeginDocument{%
  \providecommand\BibTeX{{%
    \normalfont B\kern-0.5em{\scshape i\kern-0.25em b}\kern-0.8em\TeX}}}

\acmYear{2022}\copyrightyear{2022}
\acmConference[MORSE '22]{1st ACM Workshop on AI Empowered Mobile and Wireless Sensing}{October 21, 2022}{Sydney, NSW, Australia}
\acmBooktitle{1st ACM Workshop on AI Empowered Mobile and Wireless Sensing (MORSE '22), October 21, 2022, Sydney, NSW, Australia}
\acmPrice{15.00}
\acmDOI{10.1145/3556558.3558580}
\acmISBN{978-1-4503-9522-9/22/10}

\usepackage{algorithm}
\usepackage{algpseudocode}
\usepackage{multirow}
\usepackage[export]{adjustbox}
\usepackage{booktabs}
\usepackage{caption}
\usepackage{subcaption}
\usepackage{siunitx}
\usepackage{enumitem}

\newcolumntype{x}[1]{>{\centering\arraybackslash}p{#1}}

\begin{document}

\title{A Federated Learning-enabled Smart Street Light Monitoring Application: Benefits and Future Challenges}



\author{Diya Anand}
\affiliation{%
  \institution{Electrical and Electronic Engineering\\University of Bristol}
  \city{Bristol}
  \country{UK}
}
\email{ul18753@bristol.ac.uk}

\author{Ioannis Mavromatis}
\orcid{0000-0002-3309-132X}
\affiliation{%
  \institution{Bristol Research and Innovation Laboratory (BRIL), Toshiba Europe Ltd.}
  \city{Bristol}
  \country{UK}
}
\email{Ioannis.Mavromatis@toshiba-bril.com}

\author{Pietro Carnelli}
\orcid{0000-0002-4993-5873}
\affiliation{%
  \institution{BRIL, Toshiba Europe Ltd.}
  \city{Bristol}
  \country{UK}
}
\email{Pietro.Carnelli@toshiba-bril.com}

\author{Aftab Khan}
\orcid{0000-0002-3573-6240}
\affiliation{%
  \institution{BRIL, Toshiba Europe Ltd.}
  \city{Bristol}
  \country{UK}
}
\email{Aftab.Khan@toshiba-bril.com}





\renewcommand{\shortauthors}{Anand D., Mavromatis I., et al.}

\begin{abstract}
Data-enabled cities are recently accelerated and enhanced with automated learning for improved Smart Cities applications. In the context of an Internet of Things (IoT) ecosystem, the data communication is frequently costly, inefficient, not scalable and lacks security. Federated Learning (FL) plays a pivotal role in providing privacy-preserving and communication efficient Machine Learning (ML) frameworks. In this paper we evaluate the feasibility of FL in the context of a Smart Cities Street Light Monitoring application. FL is evaluated against benchmarks of centralised and (fully) personalised machine learning techniques for the classification task of the lampposts operation. Incorporating FL in such a scenario shows minimal performance reduction in terms of the classification task, but huge improvements in the communication cost and the privacy preserving. These outcomes strengthen FL's viability and potential for IoT applications.

\end{abstract}



\maketitle

\section{Introduction}

A Smart City is described in~\cite{ubranCities} as an urban medium using Information and Communication Technologies (ICT) to promote more efficient ordinary city operations and improve the Quality of Services (QoS) received by the citizens. The objective of a Smart City is to enhance the Quality of Life (QoL) of the citizens and improve sustainability. This is achieved by promoting the digitisation of services, automation, and the use of data for intelligent responses and decisions, while autonomously adapting to different needs~\cite{sustainableSmartCities}. 

The realm of Smart Cities covers multiple areas and applications, such as Smart Transportation, Smart Urban Management, Smart Tourism, Green Cities, Smart Healthcare, etc.~\cite{smartCitiesApplications}. The technologies required for these applications are numerous but can be commonly grouped under three categories, i.e., sensing and data collection, intelligent decision support, and exchange of data and decisions between different system entities~\cite{threeKeyTechnologies}. All these technologies are part of an Internet of Things (IoT) framework~\cite{iotSmartCities} that provides the underlying infrastructure and services for their operation.  

Our work focuses on improving resource utilisation within an IoT ecosystem by moving intelligent decision-making closer to the sensors and the ``edge''. More specifically, we evaluate the feasibility of using Federated Learning (FL) within a Smart Street Light Monitoring application context. This application is a good representation of Smart Cities as it generates a large amount of data, requires increased communication bandwidth for their exchange, and is resource-intensive when classifying whether a lamppost is operational or not.

FL~\cite{flSmartCities} is a branch of Machine Learning (ML) that relies on multiple ``clients'', e.g., edge devices, to collect and process local data for training an ML model. In turn, the client's ML model parameters are shared with a central FL server for global model aggregation. Once enough updates from clients (sample fraction per round of FL) within the network have been received, a new global model is generated and broadcast to all clients. Such methods of training local models have certain benefits important to IoT and ``smart city''-centric networks~\cite{flSmartCities}. 

IoT networks often build upon Low-power Wide-Area Network (LPWAN) wireless protocols such as LoRaWAN, Zigbee and Bluetooth~\cite{wirelessSurvey}. For example, a Smart Street Light Monitoring application can easily generate hundreds of gigabytes of data if central training and processing are required~\cite{streetLightSystem}. However, this amount of data is almost impossible to be exchanged via the low-power, low data rate IoT wireless links. In such a scenario, FL can play a pivotal role by replacing the exchange of data with the exchange of the more lightweight prediction models.

Furthermore, FL provides a variety of privacy advantages required for real-world Smart Cities applications~\cite{flPrivacy}. Data minimisation is achieved as the raw data stay on the edge device, and data leakage can be avoided in transit. For external users interacting with the system, having access to only aggregated models and data can enhance privacy. Traditional end-to-end encryption mechanisms can secure the models exchanged in transit. Finally, even when an edge device is tampered with, and the model is altered, the nature of FL and the model aggregation on the server limits individual malicious models' influence on the global output. Of course, extensions and algorithms can provide more formal guarantees such as differential privacy~\cite{differentialPrivacy}, or algorithms for concept drift detection can identify such drifts~\cite{khan2022system}.

The rest of the paper is structured as follows; Section 2 summarises the related work to optimising FL clients to their local datasets/environments. Section 3 discusses our methodology, experiments and introduces our evaluation dataset. Our experimental evaluation and results are discussed in Section 4, supported by a detailed discussion on the lessons learned from this activity. Finally, Section 5 summarises our findings, and provides suggestions for future research activities.

\section{Related Work}

ML is currently being used in various intelligent fault diagnosis methods. For example,~\cite{centralisedStreetLights,centralisedStreetLights2} present two ML-based fault detection mechanisms for street light applications. Sending the collected data to a central server, the real-time illuminance of the lamppost is evaluated, and faults are reported to the maintenance staff. The results show up to $90\%$ of fault detection accuracy for such scenarios. Such an approach increased the communication cost in the IoT system while introducing many challenges in securing the data in transit.

Personalisation in ML can enhance detection by targeting particular entities and optimising the trained models accordingly. Authors in~\cite{personalisedDeepLearning} present three ways of personalisation utilising the MNIST dataset. Their results show increased performance compared to traditional ML strategies. However, personalised models trained on a server require again increased communication overhead. Personalised models trained at the edge, even though they minimise the communication overhead, require a large dataset available for each model trained. The intermittent nature of an IoT system, where data may be scarce, can create obstacles to collecting such vast amounts of data.

FL can bridge the gap between the personalised and centralised approach. Training models locally decrease communication costs while preserving data privacy. Moreover, for edge nodes with an abundance of data, aggregating the existing models on the server side and sharing them with all edge nodes can ensure that a highly accurate model is always available for inference. However, FL can suffer from highly skewed, non-Independent, Identically Distributed (IID) data. Knowing the data types and ways of clustering them can enhance FL's accuracy. Work carried out by \cite{Zhao2018FederatedData} shows an improvement of circa $30$\% on the CIFAR-10 dataset compared to classical FedAvg using their proposed data-sharing strategy between participating clients. The authors show that sharing $5$\% of a separate global dataset across clients and initialising a model at the server on the dataset mentioned above leads to a classification performance increase. Federated Meta-Learning or FedMeta framework \cite{Chen2018FederatedCommunication} uses parameterised algorithms such as MAML and Meta-SGD to train on the client's local data and communicate the updates to the server instead of updated models in traditional FL. In FedAMP \cite{Y.Huang2021PersonalizedData}, copies of local models are kept on the cloud server with attentive message passing between clients and server leading to aggregated client models of messages passed. Both methods can again enhance FL's performance.

Whilst such techniques show general performance improvement, the practical challenges are numerous. For example, creating a separate dataset of similar distribution would require prior knowledge of client data and sharing this data with the server. This method is not aligned with FL principles of maintaining local datasets for the participating clients. Furthermore, such data-sharing techniques might not be feasible amongst thousands of sensors/edge devices in an IoT network. Similarly, sharing and storing multiple copies of clients' models requires more robust, scalable and resource-rich infrastructure and storage capabilities. This approach does not scale well in networks of hundreds of thousands of participating devices.

\section{Methodology And Experiments}\label{sec:methodology}

In this paper we investigate the feasibility of FL for a lamppost fault detection use-case and compare it against a centralised and a ``fully'' personalised approach. An FL benchmark method, using a typical averaging technique to establish a global model (inspired by \cite{McMahan2016Communication-EfficientData}), was compared to a centralised method (i.e., classical ML) and an extreme version of or a ``hyper-personalised'' FL method, whereby each lamppost uses a model trained on \textbf{only} it's dataset and was never aggregated centrally. Such ``extreme'' methods provide a good overview of their potential for accurate detection using very different training data splits.

\subsection{Convolutional Neural Network Model Selection}
\label{sec:sec03}


For most FL IoT applications, the model should be lightweight enough to train on the edge device. Moreover, FL involves broadcasting the model between clients and the server. Hence a bigger model would not meet the bandwidth limitations of FL and increase communication costs significantly. 

A Convolutional Neural Network (CNN) based on the Residual Networks (ResNet) architecture was considered ``lightweight'' enough for training and inference on IoT devices and capable of classifying incoming images. In particular, ResNets~\cite{He2015DeepRecognition} were designed to combat deep neural networks which suffer from vanishing gradients (a common problem encountered in neural network training procedures). Furthermore, such a model allowed for fair comparison amongst the three different ML strategies being investigated.

\subsection{The Dataset}
\label{sec:sec01}

\begin{figure*}[t]
    \centering
    \begin{subfigure}[b]{0.3\textwidth}
        \centering
        \includegraphics[width=\textwidth]{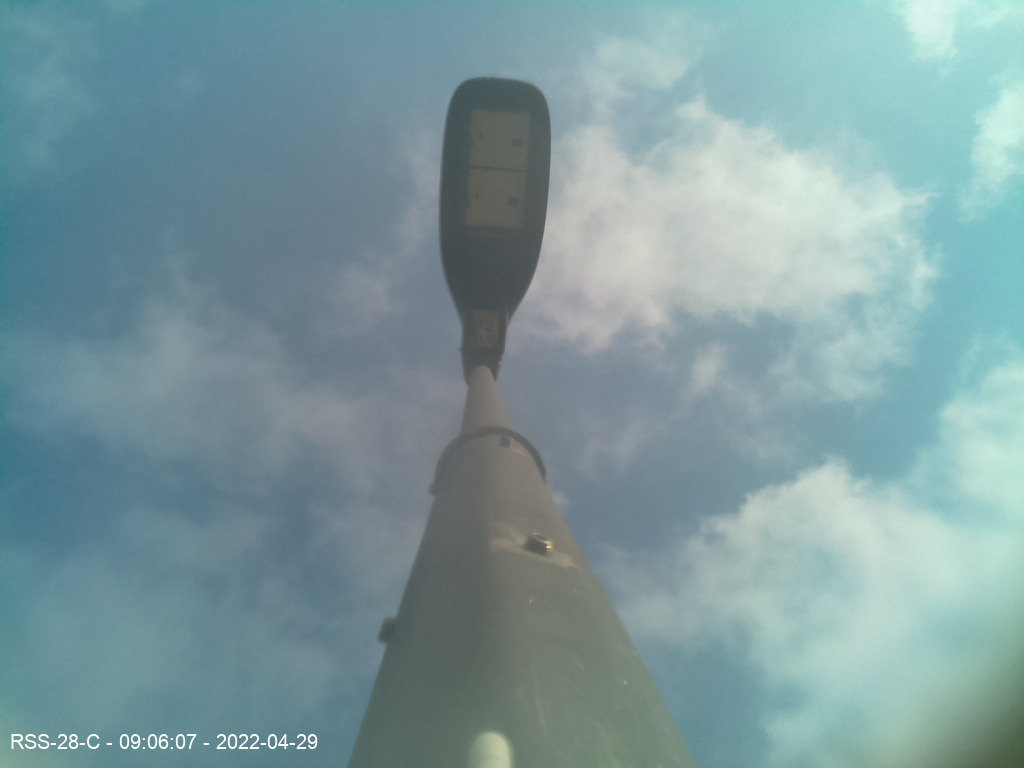}
        \caption{Node Type 0}
        \label{fig:node0}
    \end{subfigure}
    \begin{subfigure}[b]{0.3\textwidth}
        \centering
        \includegraphics[width=\textwidth]{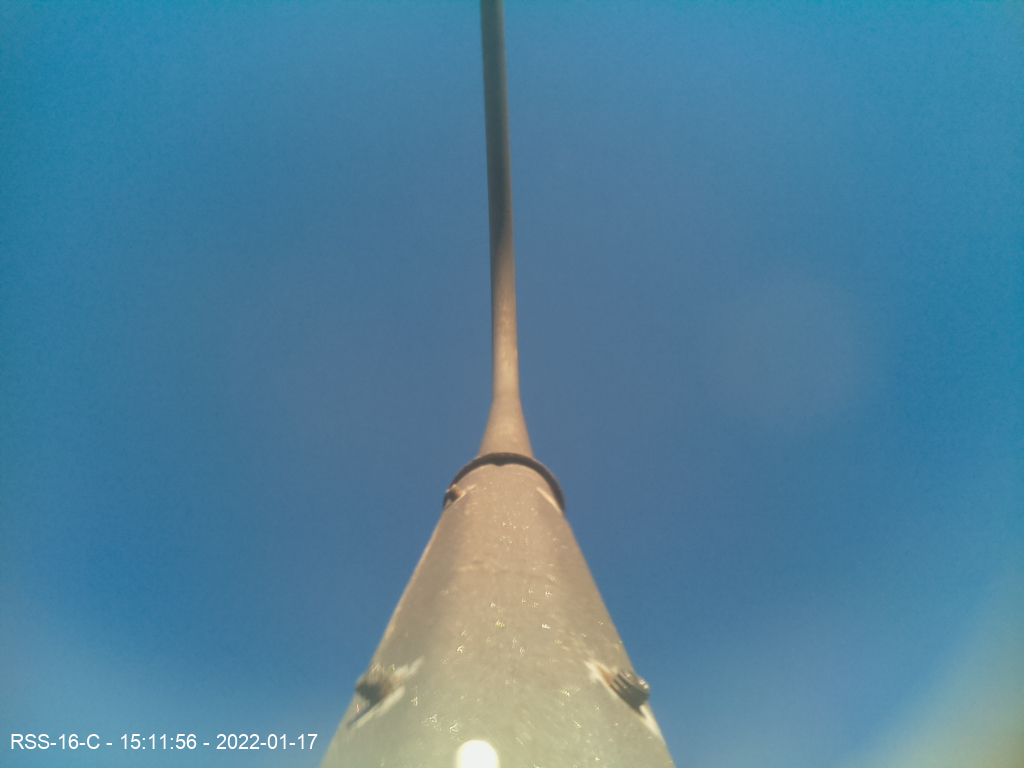}
        \caption{Node Type 1}
        \label{fig:node1}
    \end{subfigure}
    \begin{subfigure}[b]{0.3\textwidth}
        \centering
        \includegraphics[width=\textwidth]{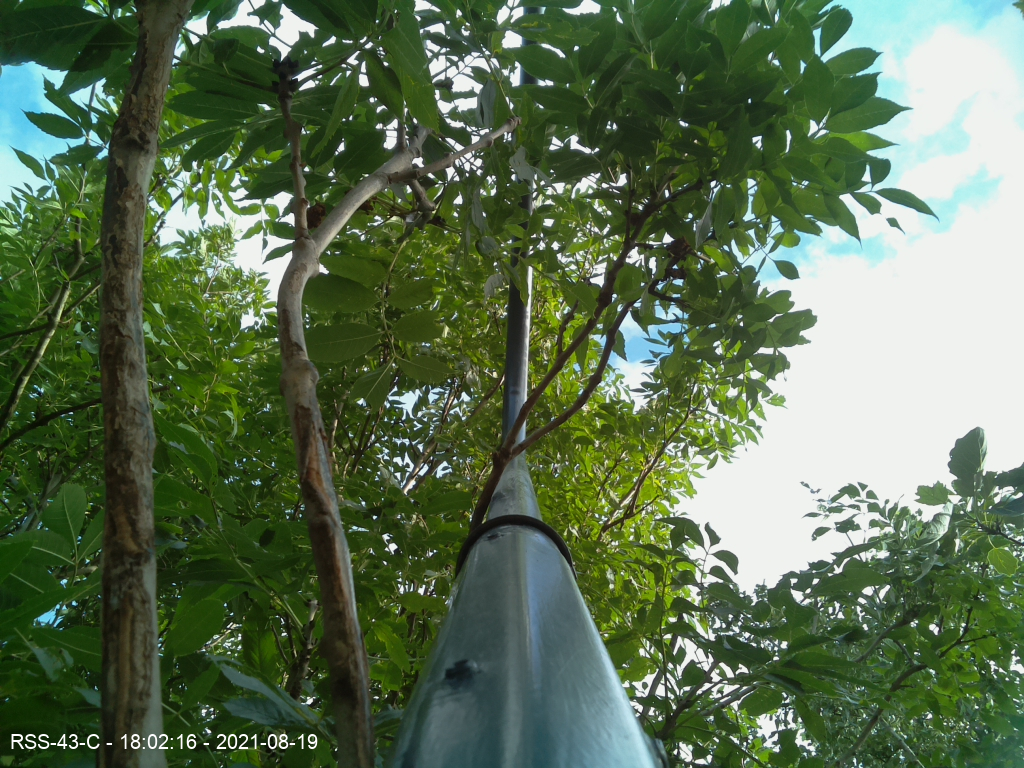}
        \caption{Node Type 2}
        \label{fig:node2}
    \end{subfigure}
    \caption{(a) Node 0: Ideal Lamppost. The light is clearly visible and this becomes easy detection of whether the light is on or off (b) Node 1: Normal Lamppost. The light is not directly visible; for detection (c) Node 2: Edge Case Lamppost. the light is completely covered by vegetation or the camera has slipped making for complicated detection.}
\label{fig:example-of-lampposts}
\end{figure*}

Our investigation is based on a large volume dataset of street light images. Since our focus is the evaluation of FL under Smart Street Light Monitoring environments, we selected the "Dataset of Images of Public Streetlights"~\cite{lamppost-dataset-paper}, generated as part of the UMBRELLA project~\cite{Farnham2021UMBRELLAPlatform}. The dataset is publicly available from Zenodo Open Repository~\cite{dataset}. The dataset consists of over $350,000$ images of streetlights, collected hourly and over a period of six months. The images come from $140$ UMBRELLA IoT nodes deployed across multiple locations in the South Gloucestershire region of the UK. The UMBRELLA nodes are currently installed at a public stretch of  $\sim$\SI{7.2}{\kilo\meter} road (about $\sim80\%$ of the nodes) and around the University of the West of England (UWE) Frenchay Campus (about $\sim20\%$ of the nodes). Since each lamppost had between $1,000 - 4,000$ images, personalised models per lamppost were possible to train and optimise for our comparison, achieving very high accuracy on each model independently.

The images in the dataset were used to determine whether the lamppost is operational or not, i.e., whether the lamppost is switched ON or OFF. The lamppost functionality is monitored during different times (once per hour), with the light expected to be ON at night and OFF during the day, as part of a partnership with the local government to ensure road safety.  As ``night'' is considered the time period from ``15-minutes before sunset'' until ``15-minutes after sunrise'' and calculated independently for each day.

The images are in JPEG format with a resolution of $1024\times768$ pixels. The entries in the dataset are already pre-labelled. What is more, the dataset spans a large geographical area and various different lamppost designs, heights and operational modes. Several streetlights are partially obstructed by vegetation or are outside the Field of View (FoV) of the camera. Finally, the cameras facing the sky are susceptible to weather conditions (e.g., rain, snow, direct sunlight, etc.) that can partially or entirely alter the quality of the images taken.  All the above generate ``interesting'' and unique edge-cases when evaluating FL within the context of a Smart Street Light Monitoring application.

\subsection{Node Categorisation}
Figure~\ref{fig:example-of-lampposts} shows some example images of the dataset used. Our evaluation and discussion are based on further grouping the nodes in the three categories seen in Figure~\ref{fig:example-of-lampposts}. As discussed in Section~\ref{sec:exp_evaluation}, we grouped the nodes into three categories with respect to the Line-of-Sight (LoS) to the lamppost (being inside the camera's FoV or not) and whether there is any obstruction by vegetation. 

More specifically, Figure~\ref{fig:node0} is labelled as ``node type 0'', modelling an ideal lamppost image; it possesses a clear view of the lamppost light, making the binary classification task of whether the light is on or off relatively simple. Figure~\ref{fig:node1} is an intermediate case labelled as ``node type 1''; due to the positioning of the UMBRELLA node, only the pole of the lamppost is visible. For this node type, the ``turned-off'' and ``turned-on'' lampposts can be classified by a human looking at the luminance of an image. However, depending on the camera's position inside the node and the weather conditions, the classification is not always easy with bare eyes. Finally, Figure \ref{fig:node2}, referred to as ``node type 2'', represents the most challenging type captured; this subclass consists of images with no view of the lamppost due to vegetation or the camera being mispositioned. The labelling of each node was done manually before the evaluation. For that, we considered the unique characteristics of each node. The labelling is later used during our evaluation process (fed as a CSV file in our algorithm)\footnote{A copy of this file can be found in the following link:\\ \href{https://www.dropbox.com/s/ydxioouluet3gwf/nodetypes.csv?dl=0}{https://www.dropbox.com/s/ydxioouluet3gwf/nodetypes.csv?dl=0}}.

\subsection{Data Pre-Processing}
\label{sec:sec02}

\begin{figure}[h]
	\includegraphics[width=1\columnwidth]{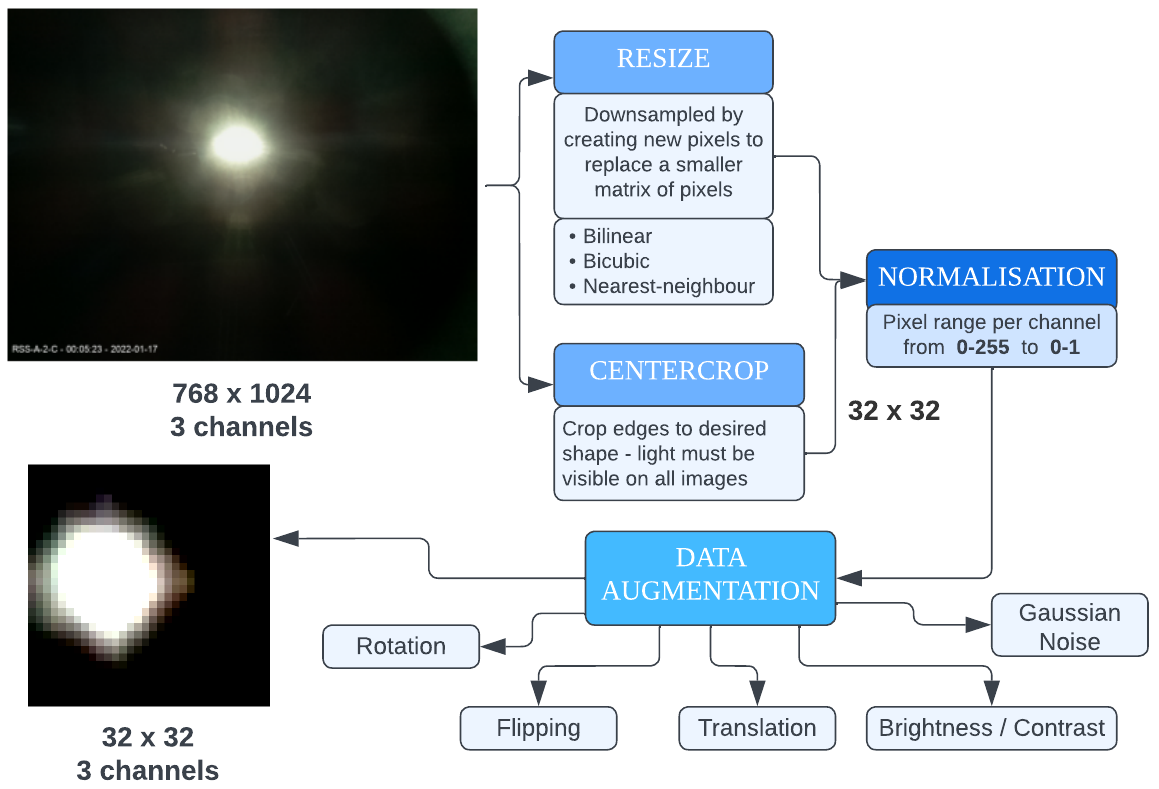}
	\caption{Flowchart of Data Pre-Processing Pipeline}
	\label{fig:PFL} 
\end{figure}

The original, `raw' lamppost dataset consists of images of $1024\times768$ Red, Green, Blue (RGB) pixels which are far too big for most CNNs trained on edge devices (such as the Nvidia Jetson Nano). To reduce the computational and memory footprint during training and deployment we reduced the images to $32\times32$ pixels with three channels (RGB representation) as shown in Figure~\ref{fig:PFL} by resizing, cropping and down-sampling the image using a bilinear interpolation method. Finally we normalised the reduced RGB images by subtracting the mean from each pixel and dividing it by the standard deviation (Figure~\ref{fig:PFL}). 

\begin{table*}[t]
    \centering
    \renewcommand{\arraystretch}{1.1}
    \caption{Overview of Experiments}

    \begin{tabular}{lccccc}
    \toprule
    Method & Trained on & \#Training Devices & Evaluated on & \#Testing Devices & \#Models\\
    \midrule
    \multirow{3}{*}{Personalised} & Normal Nodes  & 133 & Normal Nodes & 133 & 133\\
    & Edge-Case Nodes  & 7 & Edge-Case Nodes & 7 & 7\\
    \cmidrule{2-6}
    & All nodes or devices & 140 & All nodes or devices & 140 & 140\\
    \midrule
    \multirow{3}{*}{Centralised} & All nodes or devices & 140 & Normal Nodes & 133 & 1\\
    & All nodes or devices & 140 & Edge-Case Nodes & 7 & 1\\
    \cmidrule{2-6}
    & All nodes or devices & 140 & All nodes or devices & 140 & 1 \\
    \midrule
    \multirow{3}{*}{FL benchmark} & All nodes or devices & 140 & Normal Nodes & 133 & 1\\
    & All nodes or devices & 140 & Edge-Case Nodes & 7 & 1\\
    \cmidrule{2-6}
    & All nodes or devices & 140 & All nodes or devices & 140 & 1\\
    \midrule
    
    \end{tabular}

    \label{tab:overview-experiments}

\end{table*}

\section{Experimental Evaluation}~\label{sec:exp_evaluation}
As discussed in Section~\ref{sec:methodology} our evaluation compares a centralised, a ``fully'' personalised, and an FL approach. The dataset was split into a training and testing dataset; the test set consisted of 20\% of images from every lamppost node/device, and the remaining 80\% constituted the training dataset. This was consistent throughout all experiments. For our performance investigation, we also considered the type of nodes. We combined grouped the nodes of types 0 and 1 while we separately evaluated the nodes of type 2 to observe how the performance can degrade when facing such edge cases.

The developed centralised model and method generates a single model for the entire lamppost dataset for training. The ``fully'' personalised method is demonstrated by treating each lamppost as a separate entity. A model is trained solely on its own dataset (i.e., with no FL aggregation). Finally, the FL approach is based on 140 clients (i.e., one per lamppost), aggregating their models using the FedAvg algorithm. All experiments are summarised in Table~\ref{tab:overview-experiments}.

\subsection{Results for all Training/Testing Methods}

\begin{table*}
    \centering
    \renewcommand{\arraystretch}{1.1}
    \caption{Experimental Results for all Three Implemented Methods of Training and Testing.}
    \begin{tabular}{lx{1.6cm}x{1.6cm}ccx{1.8cm}ccc}
    \toprule
    Method & \#training lampposts & \#test lampposts & \#FL clients & \#Models & \#Training samples & \#Test samples & Accuracy (\%) & F1-Score \\
    \midrule
    \multirow{3}{*}{Personalised} & 133\textsuperscript{\textdagger} & 133\textsuperscript{\textdagger} & -- & 133 & 281804 & 70549 & 98.57 & 0.990\\
    & 7\textsuperscript{$\ddagger$} & 7\textsuperscript{$\ddagger$} & -- & 7 & 5891 & 1476 & 94.82 & 0.945 \\
    \cmidrule{2-9}
    & 140 & 140 & -- & 140 & 287695 & 72025 & 98.25 & 0.984 \\
    \midrule
    \multirow{3}{*}{Centralised} & 140 & 133\textsuperscript{\textdagger} & -- & 1 & 287695 & 70549 & 98.41 & 0.988\\
    & 140 & 7\textsuperscript{$\ddagger$} & -- & 1 & 287695 & 1476 & 93.39 & 0.943 \\
    \cmidrule{2-9}
    & 140 & 140 & -- & 1 & 287695 & 72025 & 98.01 & 0.983 \\
    \midrule
    \multirow{3}{*}{FL benchmark} & 140 & 133\textsuperscript{\textdagger} & 140 & 1 & 287695 & 70549 & 95.89 & 0.967\\
    & 140 & 7\textsuperscript{$\ddagger$} & 140 & 1 & 287695 & 1476 & 92.15 & 0.932 \\
    \cmidrule{2-9}
    & $\mu$ & 140 & -- & -- & 287695 & 72025 & 94.02 & 0.949 \\
    \bottomrule\\
    \multicolumn{8}{l}{\textsuperscript{\textdagger}actual normal case}\\
    \multicolumn{8}{l}{\textsuperscript{$\ddagger$}actual edge case}\\
    \multicolumn{8}{l}{$\mu$ average of models}\\
    \end{tabular}
    \label{tab:final_results}

\end{table*}

\begin{figure*}[t]
	\includegraphics[width=0.8\textwidth]{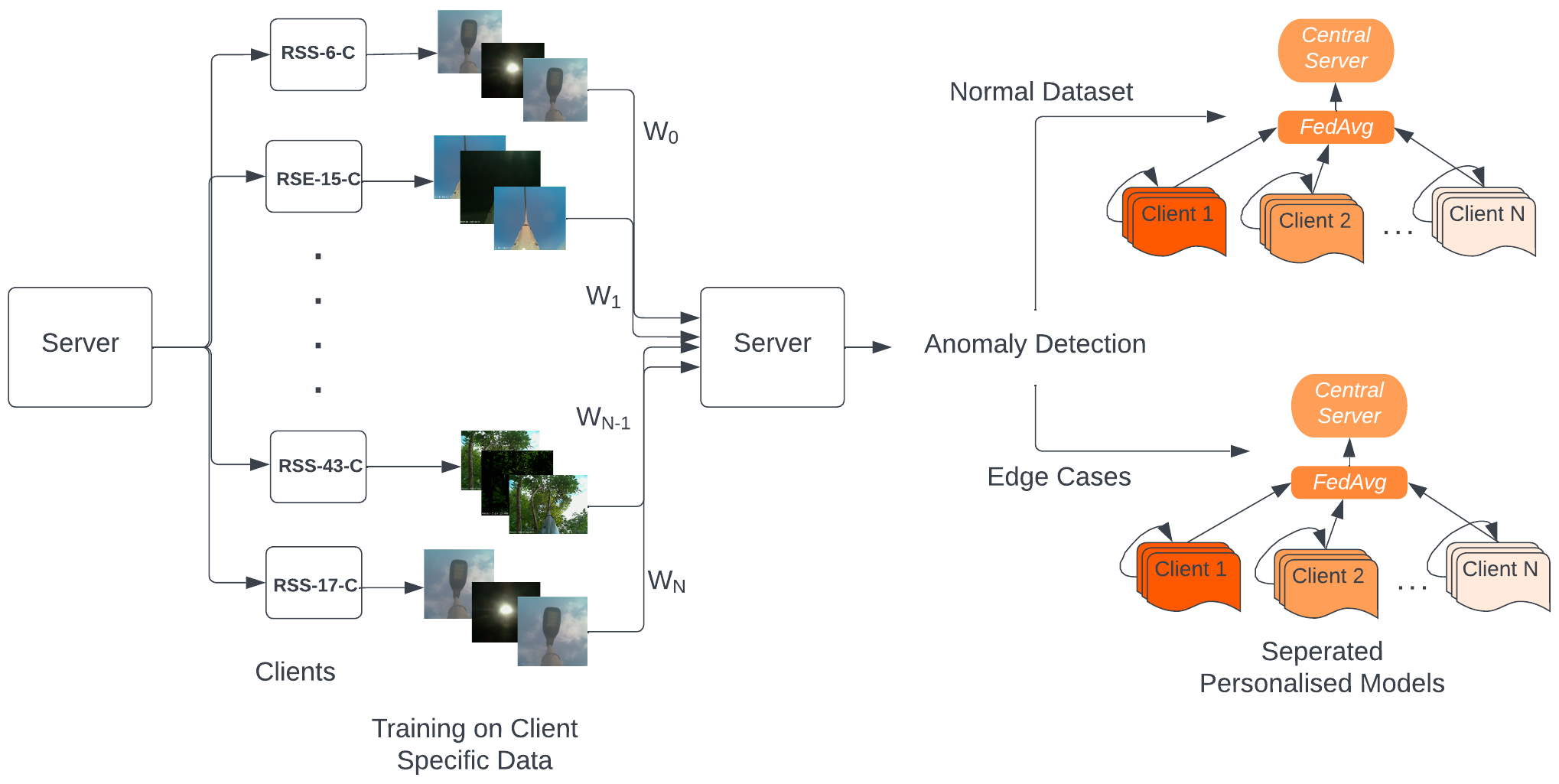}
	\caption{Flowchart of Proposed Approach for Personalised Federated Learning}
	\label{fig:our-FL-overview-method}
\end{figure*}


Our results are summarised in Table~\ref{tab:final_results}. Considering the ``normal'' nodes, as expected, the fully personalised and centralised methods generated higher accuracy ($98.57\%$ and $98.41\%$ respectively) and F1-scores ($0.99$ and $0.988$ respectively) than the benchmark FL method (accuracy of $96.7\%$ and F1-score of $0.967$). Both methods were allowed to train until fully converged (with minimal overfitting).

The difference is less prominent when the ``edge'' nodes (node type 2) are considered. Again, personalised and centralised methods slightly outperform FL, but only by a couple of percentage points. The above results are due to having either access to the entire dataset with a single model to train (centralised) or having each client train on a single lamppost's datasets (fully personalised method). 


\subsection{Discussion and Observations}~\label{subsec:discussion}

Whilst the personalised method fractionally outperformed the centralised and FL training methods, it still achieved a lower accuracy and F1-score than required for immediate deployment. Considering a city-scale deployment, such a system will still produce false positives/negatives at a rate not easily monitored by local government officials. Even at the fairly limited coverage provided within our dataset (140 nodes and lampposts), an error rate of $1.75\%$ will result in tens if not more daily alerts. Given the diversity of the node types and camera locations, more sophisticated classification algorithms are required for better accuracy. For example, taking into account multiple concurrent decisions can reduce the error rate as falsely classified results will be the minority of the reported values.

The ``fully'' personalised method relies on having a large enough amount of data stored locally on the lamppost edge device for individual training. This constitutes a significant problem when considering the resource-constrained nature of the current IoT devices or when ``new'' lampposts join the network (as there is no available global model).

Considering the centralised classification, as seen, it performs almost as well as the fully personalised method. However, it relies on having all the lamppost data transferred to a central server for processing and training. Currently, the entire compressed lamppost dataset is circa $150$GB in volume. Admittedly, this is without any of our pre-processing and dimensionality reduction techniques applied to the images. However, this is still a good indication of the costly nature of exchanging such a large volume of data. Furthermore, this will also result in privacy and confidentiality issues that may arise with the data being in transit.

The benchmark FL performed the worst out of the three evaluated. This is likely due to the combination and equal weighting during aggregation of the edge case nodes (i.e., type 2). The FL method was several percentage points lower in accuracy and F1-score than the personalised and centralised methods. However, such a method provides not only increased data privacy but also a global model too, ready to be used for initialising training or inference on a new lamppost/edge device joining the network. Therefore, gaining some scalability advantage and drastically reducing the overall data communication volume during training.

\section{Conclusions and Future Research}

In this paper, we evaluated three different methods of detecting lamppost operability in a smart city environment. As seen, a ``fully'' personalised method provides a strong performance but does not scale well. On the other hand, a centralised approach is very demanding on the communication overhead introduced. FL can provide benefits of both worlds but still lacks in terms of accuracy. We suggest the following for future FL and personalised FL research challenges:

\begin{enumerate}[label=(\roman*)]
    \item  We are currently experimenting with a ``tuned'' personalised method, whereby certain model layers are trained and optimised solely on each device and dataset. In contrast, the remaining model layers are used for aggregation when receiving updates from the global FL parameter server. Work is still ongoing, with promising results regarding accuracy observed and the reduction of communication costs.
    \item  Clustered personalised FL based on client model parameters received at the FL parameter server (proposed method shown in Figure~\ref{fig:our-FL-overview-method}). Ideally, this would allow for early detection of the lamppost node types, which could then be separately aggregated into multiple (but concurrent) FL global models. Any new connected lamppost joining the network could receive a copy of each FL global model and run local tests to evaluate performance on its local dataset before conducting local training/tuning optimisation. Early experiments suggest it is possible to detect and classify extreme edge case nodes (i.e., node type 2) fairly accurately, but we fail to detect the other node types. Whilst having only two global models for FL would work, ideally, we want to personalise the clusters to the extent that we achieve even higher accuracies, again to drastically reduce notifications/alerts received by the government officials monitoring the system and prioritising lampposts for maintenance.
    \item Given the hypothesised upper bound of performance (currently, the personalised method achieved an averaged F1-score of 0.98), it may suggest that further FL personalisation strategies may struggle to gain significant improvements. As such, in particular with ML, often the dataset might be limiting in terms of achieving such a high performance; for example details/nuances might be missed or averaged out when the images are drastically reduced in size during our pre-processing step (see Figure \ref{fig:PFL}). Consequently, we have been experimenting with image metadata and other image statistics, particularly the mean and median green pixel values. Our intuition is that in extreme edge cases, where the lighting element is not visible to the monitoring sensor camera, small amounts of reflected light (or remnants of refracted light through vegetation) might be detectable. Our preliminary experiments using such metadata to improve detection accuracy have proven relatively successful but need careful calibration and integration into a robust and scalable personalised FL method.
    \item Communication overhead in FL can be further improved by introducing selective update strategies, such as dynamic sampling or selective masking on the models exchanged~\cite{anwar2022methodsSU,9546691}. Such methods can enhance the system's scalability, mainly when introducing thousands of clients. Furthermore,  adaptive compression on the generated models~\cite{anwar2022systemAC,8171350} can bring even more benefits by reducing the communication overhead and enabling the exchange of data over low data rate IoT technologies and longer distances.
\end{enumerate}

\begin{acks}
This work is funded in part by Toshiba Europe Ltd. UMBRELLA project is funded in conjunction with South Gloucestershire Council by the West of England Local Enterprise Partnership through the Local Growth Fund, administered by the West of England Combined Authority.
\end{acks}

\bibliographystyle{ACM-Reference-Format}
\bibliography{refs}


\begin{thebibliography}{26}


\ifx \showCODEN    \undefined \def \showCODEN     #1{\unskip}     \fi
\ifx \showDOI      \undefined \def \showDOI       #1{#1}\fi
\ifx \showISBNx    \undefined \def \showISBNx     #1{\unskip}     \fi
\ifx \showISBNxiii \undefined \def \showISBNxiii  #1{\unskip}     \fi
\ifx \showISSN     \undefined \def \showISSN      #1{\unskip}     \fi
\ifx \showLCCN     \undefined \def \showLCCN      #1{\unskip}     \fi
\ifx \shownote     \undefined \def \shownote      #1{#1}          \fi
\ifx \showarticletitle \undefined \def \showarticletitle #1{#1}   \fi
\ifx \showURL      \undefined \def \showURL       {\relax}        \fi
\providecommand\bibfield[2]{#2}
\providecommand\bibinfo[2]{#2}
\providecommand\natexlab[1]{#1}
\providecommand\showeprint[2][]{arXiv:#2}

\bibitem[\protect\citeauthoryear{Anwar, Carnelli, and Khan}{Anwar
  et~al\mbox{.}}{2022a}]%
        {anwar2022methodsSU}
\bibfield{author}{\bibinfo{person}{Saif Anwar}, \bibinfo{person}{Pietro~E
  Carnelli}, {and} \bibinfo{person}{Aftab Khan}.}
  \bibinfo{year}{2022}\natexlab{a}.
\newblock \bibinfo{title}{{Methods and Systems for Remote Training of a Machine
  Learning Model}}.
\newblock
\newblock
\newblock
\shownote{US Patent App. 16/952,308.}


\bibitem[\protect\citeauthoryear{Anwar, Carnelli, and Khan}{Anwar
  et~al\mbox{.}}{2022b}]%
        {anwar2022systemAC}
\bibfield{author}{\bibinfo{person}{Saif Anwar}, \bibinfo{person}{Pietro~E
  Carnelli}, {and} \bibinfo{person}{Aftab Khan}.}
  \bibinfo{year}{2022}\natexlab{b}.
\newblock \bibinfo{title}{{System and Method for Adaptive Compression in
  Federated Learning}}.
\newblock
\newblock
\newblock
\shownote{US Patent App. 16/952,705.}


\bibitem[\protect\citeauthoryear{Chen, Luo, Dong, Li, and He}{Chen
  et~al\mbox{.}}{2018}]%
        {Chen2018FederatedCommunication}
\bibfield{author}{\bibinfo{person}{F. Chen}, \bibinfo{person}{M. Luo},
  \bibinfo{person}{Z. Dong}, \bibinfo{person}{Z. Li}, {and} \bibinfo{person}{X.
  He}.} \bibinfo{year}{2018}\natexlab{}.
\newblock \showarticletitle{{Federated meta-learning with fast convergence and
  efficient communication}}.
\newblock \bibinfo{journal}{\emph{arXiv preprint}} (\bibinfo{year}{2018}).
\newblock


\bibitem[\protect\citeauthoryear{Ding, Nemati, Ranaweera, and Choi}{Ding
  et~al\mbox{.}}{2020}]%
        {wirelessSurvey}
\bibfield{author}{\bibinfo{person}{Jie Ding}, \bibinfo{person}{Mahyar Nemati},
  \bibinfo{person}{Chathurika Ranaweera}, {and} \bibinfo{person}{Jinho Choi}.}
  \bibinfo{year}{2020}\natexlab{}.
\newblock \showarticletitle{{IoT Connectivity Technologies and Applications: A
  Survey}}.
\newblock \bibinfo{journal}{\emph{IEEE Access}}  \bibinfo{volume}{8}
  (\bibinfo{year}{2020}), \bibinfo{pages}{67646--67673}.
\newblock
\urldef\tempurl%
\url{https://doi.org/10.1109/ACCESS.2020.2985932}
\showDOI{\tempurl}


\bibitem[\protect\citeauthoryear{Farnham, Jones, Aijaz, Jin, Mavromatis, Raza,
  Portelli, Stanoev, and Sooriyabandara}{Farnham et~al\mbox{.}}{2021}]%
        {Farnham2021UMBRELLAPlatform}
\bibfield{author}{\bibinfo{person}{Tim Farnham}, \bibinfo{person}{Simon Jones},
  \bibinfo{person}{Adnan Aijaz}, \bibinfo{person}{Yichao Jin},
  \bibinfo{person}{Ioannis Mavromatis}, \bibinfo{person}{Usman Raza},
  \bibinfo{person}{Anthony Portelli}, \bibinfo{person}{Aleksandar Stanoev},
  {and} \bibinfo{person}{Mahesh Sooriyabandara}.}
  \bibinfo{year}{2021}\natexlab{}.
\newblock \showarticletitle{{UMBRELLA Collaborative Robotics Testbed and IoT
  Platform}}. In \bibinfo{booktitle}{\emph{2021 IEEE 18th Annual Consumer
  Communications and Networking Conference (CCNC)}}. \bibinfo{publisher}{IEEE},
  \bibinfo{pages}{1--7}.
\newblock
\showISBNx{978-1-7281-9794-4}
\urldef\tempurl%
\url{https://doi.org/10.1109/CCNC49032.2021.9369615}
\showDOI{\tempurl}


\bibitem[\protect\citeauthoryear{Hardy, Merrer, and Sericola}{Hardy
  et~al\mbox{.}}{2017}]%
        {8171350}
\bibfield{author}{\bibinfo{person}{C. Hardy}, \bibinfo{person}{E.~Le Merrer},
  {and} \bibinfo{person}{B. Sericola}.} \bibinfo{year}{2017}\natexlab{}.
\newblock \showarticletitle{{Distributed Deep Learning on Edge-Devices:
  Feasibility via Adaptive Compression}}. In \bibinfo{booktitle}{\emph{2017
  IEEE 16th International Symposium on Network Computing and Applications
  (NCA)}}. \bibinfo{publisher}{IEEE Computer Society}, \bibinfo{address}{Los
  Alamitos, CA, USA}, \bibinfo{pages}{1--8}.
\newblock
\urldef\tempurl%
\url{https://doi.org/10.1109/NCA.2017.8171350}
\showDOI{\tempurl}


\bibitem[\protect\citeauthoryear{He, Zhang, Ren, and Sun}{He
  et~al\mbox{.}}{2015}]%
        {He2015DeepRecognition}
\bibfield{author}{\bibinfo{person}{Kaiming He}, \bibinfo{person}{Xiangyu
  Zhang}, \bibinfo{person}{Shaoqing Ren}, {and} \bibinfo{person}{Jian Sun}.}
  \bibinfo{year}{2015}\natexlab{}.
\newblock \showarticletitle{{Deep Residual Learning for Image Recognition}}.
\newblock  (\bibinfo{date}{12} \bibinfo{year}{2015}).
\newblock


\bibitem[\protect\citeauthoryear{Ji, Jiang, Walid, and Li}{Ji
  et~al\mbox{.}}{2022}]%
        {9546691}
\bibfield{author}{\bibinfo{person}{S. Ji}, \bibinfo{person}{W. Jiang},
  \bibinfo{person}{A. Walid}, {and} \bibinfo{person}{X. Li}.}
  \bibinfo{year}{2022}\natexlab{}.
\newblock \showarticletitle{{Dynamic Sampling and Selective Masking for
  Communication-Efficient Federated Learning}}.
\newblock \bibinfo{journal}{\emph{IEEE Intelligent Systems}}
  \bibinfo{volume}{37}, \bibinfo{number}{02} (\bibinfo{date}{mar}
  \bibinfo{year}{2022}), \bibinfo{pages}{27--34}.
\newblock
\showISSN{1941-1294}
\urldef\tempurl%
\url{https://doi.org/10.1109/MIS.2021.3114610}
\showDOI{\tempurl}


\bibitem[\protect\citeauthoryear{Khan, Carnelli, Farnham, Mavromatis, and
  Portelli}{Khan et~al\mbox{.}}{2022}]%
        {khan2022system}
\bibfield{author}{\bibinfo{person}{Aftab Khan}, \bibinfo{person}{Pietro~E
  Carnelli}, \bibinfo{person}{Timothy~David Farnham}, \bibinfo{person}{Ioannis
  Mavromatis}, {and} \bibinfo{person}{Anthony Portelli}.}
  \bibinfo{year}{2022}\natexlab{}.
\newblock \bibinfo{title}{{System and Method for Detecting and Rectifying
  Concept Drift in Federated Learning}}.
\newblock
\newblock
\newblock
\shownote{US Patent App. 17/018,923.}


\bibitem[\protect\citeauthoryear{Lee, Zhang, and Rosa}{Lee
  et~al\mbox{.}}{2019}]%
        {centralisedStreetLights2}
\bibfield{author}{\bibinfo{person}{Yanming Lee}, \bibinfo{person}{Hongyi
  Zhang}, {and} \bibinfo{person}{Jimmi Rosa}.} \bibinfo{year}{2019}\natexlab{}.
\newblock \showarticletitle{Street Lamp Fault Diagnosis System Based on Extreme
  Learning Machine}.
\newblock \bibinfo{journal}{\emph{{IOP} Conference Series: Materials Science
  and Engineering}}  \bibinfo{volume}{490} (\bibinfo{date}{apr}
  \bibinfo{year}{2019}), \bibinfo{pages}{042053}.
\newblock
\urldef\tempurl%
\url{https://doi.org/10.1088/1757-899x/490/4/042053}
\showDOI{\tempurl}


\bibitem[\protect\citeauthoryear{Lin, Liu, and Wang}{Lin et~al\mbox{.}}{2021}]%
        {flPrivacy}
\bibfield{author}{\bibinfo{person}{Hui Lin}, \bibinfo{person}{Wenxin Liu},
  {and} \bibinfo{person}{Xiaoding Wang}.} \bibinfo{year}{2021}\natexlab{}.
\newblock \showarticletitle{{A Secure Federated Learning Mechanism for Data
  Privacy Protection}}. In \bibinfo{booktitle}{\emph{2021 20th International
  Conference on Ubiquitous Computing and Communications
  (IUCC/CIT/DSCI/SmartCNS)}}. \bibinfo{pages}{25--31}.
\newblock
\urldef\tempurl%
\url{https://doi.org/10.1109/IUCC-CIT-DSCI-SmartCNS55181.2021.00019}
\showDOI{\tempurl}


\bibitem[\protect\citeauthoryear{Marsal-Llacuna, Colomer-Llinàs, and
  Meléndez-Frigola}{Marsal-Llacuna et~al\mbox{.}}{2015}]%
        {ubranCities}
\bibfield{author}{\bibinfo{person}{M-Lluïsa Marsal-Llacuna},
  \bibinfo{person}{J. Colomer-Llinàs}, {and} \bibinfo{person}{J.
  Meléndez-Frigola}.} \bibinfo{year}{2015}\natexlab{}.
\newblock \showarticletitle{Lessons in urban monitoring taken from sustainable
  and livable cities to better address the Smart Cities initiative}.
\newblock \bibinfo{journal}{\emph{Technological Forecasting and Social Change}}
   \bibinfo{volume}{90} (\bibinfo{year}{2015}), \bibinfo{pages}{611--622}.
\newblock


\bibitem[\protect\citeauthoryear{Mavromatis, Stanoev, Carnelli, Jin,
  Sooriyabandara, and Khan}{Mavromatis et~al\mbox{.}}{2022a}]%
        {lamppost-dataset-paper}
\bibfield{author}{\bibinfo{person}{Ioannis Mavromatis},
  \bibinfo{person}{Aleksandar Stanoev}, \bibinfo{person}{Pietro Carnelli},
  \bibinfo{person}{Yichao Jin}, \bibinfo{person}{Mahesh Sooriyabandara}, {and}
  \bibinfo{person}{Aftab Khan}.} \bibinfo{year}{2022}\natexlab{a}.
\newblock \bibinfo{title}{A Dataset of Images of Public Streetlights with
  Operational Monitoring using Computer Vision Techniques}.
\newblock
\newblock
\urldef\tempurl%
\url{https://doi.org/10.48550/ARXIV.2203.16915}
\showDOI{\tempurl}


\bibitem[\protect\citeauthoryear{Mavromatis, Stanoev, Carnelli, Jin,
  Sooriyabandara, and Khan}{Mavromatis et~al\mbox{.}}{2022b}]%
        {dataset}
\bibfield{author}{\bibinfo{person}{Ioannis Mavromatis},
  \bibinfo{person}{Aleksandar Stanoev}, \bibinfo{person}{Pietro Carnelli},
  \bibinfo{person}{Yichao Jin}, \bibinfo{person}{Mahesh Sooriyabandara}, {and}
  \bibinfo{person}{Aftab Khan}.} \bibinfo{year}{2022}\natexlab{b}.
\newblock \bibinfo{booktitle}{\emph{{Images of Public Streetlights with
  Operational Monitoring using Computer Vision Techniques}}}.
\newblock
\urldef\tempurl%
\url{https://doi.org/10.5281/zenodo.6046759}
\showDOI{\tempurl}


\bibitem[\protect\citeauthoryear{McMahan, Moore, Ramage, Hampson, and
  Arcas}{McMahan et~al\mbox{.}}{2016}]%
        {McMahan2016Communication-EfficientData}
\bibfield{author}{\bibinfo{person}{H.~Brendan McMahan}, \bibinfo{person}{Eider
  Moore}, \bibinfo{person}{Daniel Ramage}, \bibinfo{person}{Seth Hampson},
  {and} \bibinfo{person}{Blaise Agüera~y Arcas}.}
  \bibinfo{year}{2016}\natexlab{}.
\newblock \showarticletitle{{Communication-Efficient Learning of Deep Networks
  from Decentralized Data}}.
\newblock  (\bibinfo{date}{2} \bibinfo{year}{2016}).
\newblock


\bibitem[\protect\citeauthoryear{Ouadrhiri and Abdelhadi}{Ouadrhiri and
  Abdelhadi}{2022}]%
        {differentialPrivacy}
\bibfield{author}{\bibinfo{person}{Ahmed~El Ouadrhiri} {and}
  \bibinfo{person}{Ahmed Abdelhadi}.} \bibinfo{year}{2022}\natexlab{}.
\newblock \showarticletitle{{Differential Privacy for Deep and Federated
  Learning: A Survey}}.
\newblock \bibinfo{journal}{\emph{IEEE Access}}  \bibinfo{volume}{10}
  (\bibinfo{year}{2022}), \bibinfo{pages}{22359--22380}.
\newblock
\urldef\tempurl%
\url{https://doi.org/10.1109/ACCESS.2022.3151670}
\showDOI{\tempurl}


\bibitem[\protect\citeauthoryear{Schneider and Vlachos}{Schneider and
  Vlachos}{2021}]%
        {personalisedDeepLearning}
\bibfield{author}{\bibinfo{person}{Johannes Schneider} {and}
  \bibinfo{person}{Michalis Vlachos}.} \bibinfo{year}{2021}\natexlab{}.
\newblock \showarticletitle{Personalization of Deep Learning}. In
  \bibinfo{booktitle}{\emph{Data Science -- Analytics and Applications}},
  \bibfield{editor}{\bibinfo{person}{Peter Haber}, \bibinfo{person}{Thomas
  Lampoltshammer}, \bibinfo{person}{Manfred Mayr}, {and}
  \bibinfo{person}{Kathrin Plankensteiner}} (Eds.).
  \bibinfo{publisher}{Springer Fachmedien Wiesbaden},
  \bibinfo{address}{Wiesbaden}, \bibinfo{pages}{89--96}.
\newblock


\bibitem[\protect\citeauthoryear{Silva, Khan, and Han}{Silva
  et~al\mbox{.}}{2018}]%
        {sustainableSmartCities}
\bibfield{author}{\bibinfo{person}{Bhagya~Nathali Silva},
  \bibinfo{person}{Murad Khan}, {and} \bibinfo{person}{Kijun Han}.}
  \bibinfo{year}{2018}\natexlab{}.
\newblock \showarticletitle{{Towards Sustainable Smart Cities: A Review of
  Trends, Architectures, Components, and Open Challenges in Smart Cities}}.
\newblock \bibinfo{journal}{\emph{Sustainable Cities and Society}}
  \bibinfo{volume}{38} (\bibinfo{year}{2018}), \bibinfo{pages}{697--713}.
\newblock
\showISSN{2210-6707}
\urldef\tempurl%
\url{https://doi.org/10.1016/j.scs.2018.01.053}
\showDOI{\tempurl}


\bibitem[\protect\citeauthoryear{Su, Li, and Fu}{Su et~al\mbox{.}}{2011}]%
        {smartCitiesApplications}
\bibfield{author}{\bibinfo{person}{Kehua Su}, \bibinfo{person}{Jie Li}, {and}
  \bibinfo{person}{Hongbo Fu}.} \bibinfo{year}{2011}\natexlab{}.
\newblock \showarticletitle{{Smart City and The Applications}}. In
  \bibinfo{booktitle}{\emph{2011 International Conference on Electronics,
  Communications and Control (ICECC)}}. \bibinfo{pages}{1028--1031}.
\newblock
\urldef\tempurl%
\url{https://doi.org/10.1109/ICECC.2011.6066743}
\showDOI{\tempurl}


\bibitem[\protect\citeauthoryear{Tang, Ding, Deng, Zhang, Wang, and Lv}{Tang
  et~al\mbox{.}}{2021}]%
        {centralisedStreetLights}
\bibfield{author}{\bibinfo{person}{Dongjun Tang}, \bibinfo{person}{Fulai Ding},
  \bibinfo{person}{Boya Deng}, \bibinfo{person}{Pingkang Zhang},
  \bibinfo{person}{Qingming Wang}, {and} \bibinfo{person}{Hui Lv}.}
  \bibinfo{year}{2021}\natexlab{}.
\newblock \showarticletitle{{An Intelligent Fault Diagnosis Method for Street
  Lamps}}. In \bibinfo{booktitle}{\emph{2021 International Conference on
  Internet, Education and Information Technology (IEIT)}}.
  \bibinfo{pages}{300--303}.
\newblock
\urldef\tempurl%
\url{https://doi.org/10.1109/IEIT53597.2021.00073}
\showDOI{\tempurl}


\bibitem[\protect\citeauthoryear{Umamaheswari, Priya, and Kumar}{Umamaheswari
  et~al\mbox{.}}{2021}]%
        {threeKeyTechnologies}
\bibfield{author}{\bibinfo{person}{S. Umamaheswari}, \bibinfo{person}{K.Hari
  Priya}, {and} \bibinfo{person}{S.Arun Kumar}.}
  \bibinfo{year}{2021}\natexlab{}.
\newblock \showarticletitle{{Technologies Used in Smart City Applications –
  An Overview}}. In \bibinfo{booktitle}{\emph{2021 International Conference on
  Advancements in Electrical, Electronics, Communication, Computing and
  Automation (ICAECA)}}. \bibinfo{pages}{1--6}.
\newblock
\urldef\tempurl%
\url{https://doi.org/10.1109/ICAECA52838.2021.9675707}
\showDOI{\tempurl}


\bibitem[\protect\citeauthoryear{{Y. Huang}, {L. Chu}, {Z. Zhou}, {L. Wang},
  {J. Liu}, {J. Pei}, and {Y. Zhang}}{{Y. Huang} et~al\mbox{.}}{2021}]%
        {Y.Huang2021PersonalizedData}
\bibfield{author}{\bibinfo{person}{{Y. Huang}}, \bibinfo{person}{{L. Chu}},
  \bibinfo{person}{{Z. Zhou}}, \bibinfo{person}{{L. Wang}},
  \bibinfo{person}{{J. Liu}}, \bibinfo{person}{{J. Pei}}, {and}
  \bibinfo{person}{{Y. Zhang}}.} \bibinfo{year}{2021}\natexlab{}.
\newblock \showarticletitle{{Personalized cross-silo federated learning on
  non-iid data}}.
\newblock \bibinfo{journal}{\emph{Association for the Advancement of Artificial
  Intelligence (AAAI)}} (\bibinfo{year}{2021}).
\newblock


\bibitem[\protect\citeauthoryear{Yang, Lee, Chen, Yang, Huang, and Hou}{Yang
  et~al\mbox{.}}{2020}]%
        {streetLightSystem}
\bibfield{author}{\bibinfo{person}{Yu-Sheng Yang}, \bibinfo{person}{Shih-Hsiung
  Lee}, \bibinfo{person}{Guan-Sheng Chen}, \bibinfo{person}{Chu-Sing Yang},
  \bibinfo{person}{Yueh-Min Huang}, {and} \bibinfo{person}{Ting-Wei Hou}.}
  \bibinfo{year}{2020}\natexlab{}.
\newblock \showarticletitle{{An Implementation of High Efficient Smart Street
  Light Management System for Smart City}}.
\newblock \bibinfo{journal}{\emph{IEEE Access}}  \bibinfo{volume}{8}
  (\bibinfo{year}{2020}), \bibinfo{pages}{38568--38585}.
\newblock
\urldef\tempurl%
\url{https://doi.org/10.1109/ACCESS.2020.2975708}
\showDOI{\tempurl}


\bibitem[\protect\citeauthoryear{Zanella, Bui, Castellani, Vangelista, and
  Zorzi}{Zanella et~al\mbox{.}}{2014}]%
        {iotSmartCities}
\bibfield{author}{\bibinfo{person}{Andrea Zanella}, \bibinfo{person}{Nicola
  Bui}, \bibinfo{person}{Angelo Castellani}, \bibinfo{person}{Lorenzo
  Vangelista}, {and} \bibinfo{person}{Michele Zorzi}.}
  \bibinfo{year}{2014}\natexlab{}.
\newblock \showarticletitle{{Internet of Things for Smart Cities}}.
\newblock \bibinfo{journal}{\emph{IEEE Internet of Things Journal}}
  \bibinfo{volume}{1}, \bibinfo{number}{1} (\bibinfo{year}{2014}),
  \bibinfo{pages}{22--32}.
\newblock
\urldef\tempurl%
\url{https://doi.org/10.1109/JIOT.2014.2306328}
\showDOI{\tempurl}


\bibitem[\protect\citeauthoryear{Zhao, Li, Lai, Suda, Civin, and Chandra}{Zhao
  et~al\mbox{.}}{2018}]%
        {Zhao2018FederatedData}
\bibfield{author}{\bibinfo{person}{Yue Zhao}, \bibinfo{person}{Meng Li},
  \bibinfo{person}{Liangzhen Lai}, \bibinfo{person}{Naveen Suda},
  \bibinfo{person}{Damon Civin}, {and} \bibinfo{person}{Vikas Chandra}.}
  \bibinfo{year}{2018}\natexlab{}.
\newblock \showarticletitle{{Federated Learning with Non-IID Data}}.
\newblock  (\bibinfo{date}{6} \bibinfo{year}{2018}).
\newblock


\bibitem[\protect\citeauthoryear{Zheng, Zhou, Sun, Wang, Liu, and Li}{Zheng
  et~al\mbox{.}}{2022}]%
        {flSmartCities}
\bibfield{author}{\bibinfo{person}{Zhaohua Zheng}, \bibinfo{person}{Yize Zhou},
  \bibinfo{person}{Yilong Sun}, \bibinfo{person}{Zhang Wang},
  \bibinfo{person}{Boyi Liu}, {and} \bibinfo{person}{Keqiu Li}.}
  \bibinfo{year}{2022}\natexlab{}.
\newblock \showarticletitle{{Applications of Federated Learning in Smart
  Cities: Recent Advances, Taxonomy, and Open Challenges}}.
\newblock \bibinfo{journal}{\emph{Connection Science}} \bibinfo{volume}{34},
  \bibinfo{number}{1} (\bibinfo{year}{2022}), \bibinfo{pages}{1--28}.
\newblock
\urldef\tempurl%
\url{https://doi.org/10.1080/09540091.2021.1936455}
\showDOI{\tempurl}


\end{thebibliography}

\end{document}